%% file: anonymous-submission-latex-2026.tex
\documentclass[letterpaper]{article} 
\usepackage[]{aaai2026}  
\usepackage{times}  
\usepackage{helvet}  
\usepackage{courier}  
\usepackage[hyphens]{url}  
\usepackage{graphicx} 
\usepackage{natbib}  
\usepackage{caption} 
\usepackage{algorithm}
\usepackage{algorithmic}

\usepackage{newfloat}
\usepackage{listings}
\DeclareCaptionStyle{ruled}{labelfont=normalfont,labelsep=colon,strut=off} 
\floatstyle{ruled}
\newfloat{listing}{tb}{lst}{}
\floatname{listing}{Listing}
\usepackage{amssymb}
\usepackage{amsmath}
\usepackage{siunitx}
\usepackage{booktabs}
\usepackage{multirow}
\usepackage{bm}
\usepackage{amsthm}

\title{Magnitude-Modulated Equivariant Adapter for Parameter-Efficient Fine-Tuning of Equivariant Graph Neural Networks}
\author{
    Dian Jin\textsuperscript{\rm 1}, 
    Yancheng Yuan\textsuperscript{\rm 1}\thanks{Corresponding author.}, 
    Xiaoming Tao\textsuperscript{\rm 1}
}
\affiliations{
    \textsuperscript{\rm 1}The Hong Kong Polytechnic University\\
    diann.jin@connect.polyu.hk, yancheng.yuan@polyu.edu.hk

}

\usepackage{bibentry}

\begin{document}

\maketitle

\begin{abstract}
Pretrained equivariant graph neural networks based on spherical harmonics offer efficient and accurate alternatives to computationally expensive ab-initio methods, yet adapting them to new tasks and chemical environments still requires fine-tuning. Conventional parameter-efficient fine-tuning (PEFT) techniques, such as Adapters and LoRA, typically break symmetry, making them incompatible with those equivariant architectures. ELoRA, recently proposed, is the first equivariant PEFT method. It achieves improved parameter efficiency and performance on many benchmarks. However, the relatively high degrees of freedom it retains within each tensor order can still perturb pretrained feature distributions and ultimately degrade performance.
To address this, we present Magnitude-Modulated Equivariant Adapter (MMEA), a novel equivariant fine-tuning method which employs lightweight scalar gating to modulate feature magnitudes on a per-order and per-multiplicity basis. We demonstrate that MMEA preserves strict equivariance and, across multiple benchmarks, consistently improves energy and force predictions to state-of-the-art levels while training fewer parameters than competing approaches. These results suggest that, in many practical scenarios, modulating channel magnitudes is sufficient to adapt equivariant models to new chemical environments without breaking symmetry, pointing toward a new paradigm for equivariant PEFT design. 
\end{abstract}

\begin{links}
    \link{Code}{https://github.com/CLaSLoVe/MMEA}
\end{links}

\section{Introduction}

Density-functional theory (DFT) offers a reliable description of electronic structure, and has already become a standard technique in most branches of chemistry and materials science \cite{burke2012perspective}.  However, even with an exascale system, the scale one could achieve with DFT is limited due to its cubic scaling on system size \cite{fiedler2023predicting}. Thus, the costs of considering a huge number of configurations in long simulations, or of dealing with the exponentially growing number of possible materials in high-throughput screenings will remain prohibitive \cite{behler2017first}.

To bridge this gap, researchers have turned to deep learning-based potentials \cite{behler2007generalized}, which greatly accelerate computer simulations, while preserving quantum mechanical accuracy \cite{behler2017first}. Those methods \cite{behler2007generalized,smith2017ani,schutt2018schnet,gilmer2017neural,zhang2018deep,unke2019physnet,unke2021spookynet,aykent2025gotennet,chang2025mgnn}, hold significant promise in addressing the accuracy-efficiency trade-off and therefore allow the expansion of the spatial and temporal scales of molecular dynamics simulations \cite{wang2025elora}. 
Recent work has produced large, pretrained foundation models such as Equiformer \cite{liao2022equiformer}, NequIP \cite{batzner20223}, Uni-Mol \cite{zhou2023uni}, MACE \cite{Batatia2022mace}. Among all these models, those equivariant graph neural networks based on spherical harmonics are particularly powerful, as they are not only inherently designed to respect rotational, translational, and permutational symmetries, but also to model high‑order physical information \cite{anderson2019cormorant}. This leads to exceptional sample efficiency: chemical accuracy can often be achieved with just a few hundred to a few thousand local structures, as it is an important benefit when high-quality quantum data is costly to obtain \cite{maruf2025equivariant}.

Nonetheless, when a target system is poorly represented in the pretraining data, such as rare chemistry configurations, a foundation model can lose its accuracy. Fortunately, it has been shown in multiple domains  \cite{howard2018universal,devi2024optimal} that fine‑tuning on a small, task‑specific dataset can restore accuracy in such cases and typically yields better results than training from scratch. However, full-parameter fine-tuning risks over-fitting and catastrophic forgetting \cite{bethune2025scaling}. Parameter-efficient fine-tuning (PEFT) techniques like Low-Rank Adaptation (LoRA) \cite{hu2022lora} are attractive alternatives, yet mixing different tensor orders in an equivariant network breaks symmetry.

An equivariant extension of LoRA called ELoRA \cite{wang2025elora}, addresses this issue by introducing path-dependent low-rank adapters into each tensor channel, demonstrating superior performance compared to full-parameter fine-tuning. This clearly highlights the potential of PEFT. By restricting the degrees of freedom during learning, ELoRA reduces the risk of overfitting and catastrophic forgetting, resulting in a model with stronger generalization capabilities.

However, we observe that ELoRA still maintains relatively high degrees of freedom within each tensor order. Building on the insight that a well-trained equivariant backbone inherently provides a robust foundation for each tensor order, we propose an even lighter fine-tuning strategy that employs only per-channel scaling adjustments. This approach strictly preserves equivariance, introduces fewer additional parameters, and consistently outperforms ELoRA across diverse molecular and materials benchmarks.

Our main contributions are listed below:

\begin{enumerate}
\item We introduce MMEA, an equivariant PEFT approach for equivariant graph neural networks based on spherical harmonics that leverages dynamic magnitude modulation on a per-channel and per-multiplicity basis. 
\item Evaluations on multiple molecular benchmarks show that MMEA achieves state-of-the-art fine-tuning performance, surpassing previous methods while requiring fewer trainable parameters.
\item We integrate MMEA into the widely used e3nn framework, providing a convenient and efficient fine-tuning solution for the community. 
\end{enumerate}

\section{Related Works}

\subsection{Equvariant Graph Neural Networks}

Graph Neural Networks (GNNs), especially those based on the Message Passing Neural Network (MPNN) framework \cite{gilmer2017neural}, have demonstrated strong performance in processing graph-structured data such as social networks \cite{liu2021content} and recommender systems \cite{wu2022graph}. In molecular applications, small molecules can naturally be represented as graphs, where atoms serve as nodes and bonds as edges. Consequently, GNNs have been widely adopted for molecular property prediction \cite{lee2023principal,wieder2020compact}. However, unlike abstract graphs, molecules possess three-dimensional structures that are crucial for determining their physical and chemical properties, such as energy and atomic forces. This motivates the integration of 3D structural information into GNN models.

Early efforts focused on incorporating invariant geometric features into GNNs to respect the rotational and translational symmetries of molecular systems. For example, {SchNet} \cite{schutt2018schnet} encodes pairwise atomic distances (i.e., two-body information) using radial basis functions. {DimeNet} \cite{gasteiger2020directional} extends this approach by modeling angular information between triplets of atoms (three-body interactions), while {GemNet} \cite{gasteiger2021gemnet} further includes torsional angles (four-body interactions). These models, often referred to as {invariant GNNs}, produce outputs that remain unchanged under global rotations or translations of the molecule. While effective, they rely on carefully designed geometric descriptors and handcrafted feature engineering, which limit scalability and make it difficult to model directional quantities such as forces.

To overcome these limitations, researchers have proposed {equivariant GNNs}, whose outputs transform predictably under geometric transformations. These models can capture both scalar and directional information and are generally more expressive. Two major classes of equivariant GNNs have emerged.

The first class includes scalar–vector models such as {PaiNN} \cite{schutt2021equivariant}, {EGNN} \cite{satorras2021n}, and {TorchMD-Net} \cite{tholke2022torchmd}. These methods maintain both scalar and three-dimensional vector features at each node, updating them jointly during message passing. They are computationally efficient and suitable for tasks involving directionally dependent quantities. However, from a theoretical perspective, these models only capture first-order tensor interactions and thus exhibiting limited fitting capability. Moreover, some studies have shown that such architectures may suffer from the ``zero function" problem \cite{cen2024high}, where outputs can collapse to zero under certain conditions, limiting their expressiveness.

The second class of methods involves higher‑order tensor models based on spherical harmonics and Clebsch–Gordan (CG) tensor products. Tensor Field Net \cite{thomas2018tensor} and Cormorant \cite{anderson2019cormorant} pioneered this strategy by decomposing features into spherical harmonic components and using CG tensor products to achieve equivariant message passing. These higher‑order tensor–based equivariant models excel primarily because they can explicitly encode higher‑order geometric and physical information, thereby capturing complex interactions that standard GNNs difficult to represent. In particular, the subsequent MACE family of foundation models \cite{Batatia2022mace} extends the architecture from traditional pairwise interactions to higher‑order many‑body interactions, significantly improving the accuracy of energy and force predictions.


\subsection{Parameter-Efficient Fine-Tuning}
PEFT methods aim to reduce the number of parameters that need to be updated during the fine-tuning process. The main goal is to address the inefficiency of full-parameter updates as model size and task complexity increase. By limiting the degrees of freedom in model updates, PEFT methods can help mitigate overfitting in certain scenarios.
PEFT is primarily applied in CV and NLP, and there are many different approaches. For example, Adapter \cite{houlsby2019parameter} tuning introduces bottleneck modules between layers to enable efficient adaptation; LoRA \cite{hu2022lora} decomposes weight matrices into low-rank representations, reducing the number of trainable parameters; BitFit \cite{zaken2021bitfit} only updates bias terms while keeping all other parameters fixed; FiLM \cite{perez2018film} modifies intermediate representations through feature-wise affine transformations, enabling task-specific modulation without full parameter updates. 
Recently, research has extended these techniques to other domains, such as GNNs, including methods like AdapterGNN \cite{li2024adaptergnn}. However, most existing approaches map inputs to bottleneck modules and then back to outputs, which mixes irreducible representations of different orders in equivariant GNNs. This breaks the equivariance property and leads to the loss of symmetry guarantees, which can negatively impact performance \cite{wang2025elora}.
To address this, ELoRA \cite{wang2025elora} is proposed, which introduces a path-dependent decomposition for weight updates in SO(3) equivariant GNNs, ensuring that low-rank adaptations respect the symmetry constraints intrinsic to the model. This not only reduces the number of trainable parameters, but also enhanced prediction performance, enhancing data efficiency.


\section{Methodology}

We propose a novel equivariant adapter, Magnitude-Modulated Equivariant Adapter, for the fine-tuning process of Equivariant GNNs. Our approach is grounded in the path-dependent paradigm, but unlike the LoRA-based ELoRA method, MMEA employs scalar gating to adjust each channel’s magnitude without mixing different multiplicities under the same irreducible order.  
Consequently, it better leverages the information in the pretrained model, mitigates overfitting, and introduces fewer additional parameters.

\begin{figure}[htbp]
    \centering
    \includegraphics[width=0.98\linewidth]{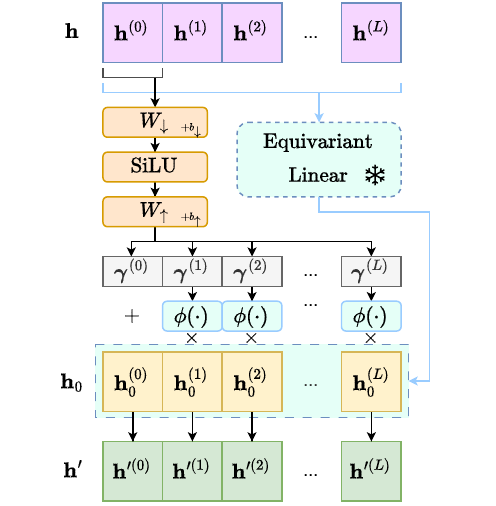} 
    \caption{Overview of the Magnitude‑Modulated Equivariant Adapter. Rounded boxes denote operators, dashed outline indicates components of the original architecture (untrainable components).}
    \label{fig:mag}
\end{figure}

The main architecture of MMEA is shown in Fig.~\ref{fig:mag}. Our proposed method is inspired by physical intuition: in a well-trained equivariant GNN model, the multiplicity channel for each order already constitutes a robust basis, and allowing them to mix freely during fine-tuning may distort the geometry of the pretrained feature space. Our method avoids mixing different multiplicities, which helps preserve the knowledge learned by the pre-trained model more effectively. Next, we introduce the details of the MMEA architecture.


\paragraph{Node feature space}
The node feature space can be denoted as a direct sum of irreducible 
representations \cite{thomas2018tensor}:
\begin{equation}
  \mathcal{H} \;:=\;
  \bigoplus_{\ell = 0}^{L} \mathcal{H}^{(\ell)},
  \qquad
  \mathcal{H}^{(\ell)} \;:=\;
  V^{(\ell)} \otimes \mathbb{R}^{1 \times m_\ell},
  \label{eq:space}
\end{equation}
where $V^{(\ell)}$ is the irreducible representation of order
$\ell$ with dimension $d_\ell = 2\ell + 1$, $m_\ell$
is its multiplicity, and $\otimes $ denotes the Kronecker product. $L$ is the maximum representation order.
The {SO}(3) (Special Orthogonal Group in 3 Dimensions)  action $g$ is
\begin{equation}
  g \cdot \bigl( v \otimes a \bigr)
  \;:=\;
  (\rho^{(\ell)}(g)\,v) \;\otimes\; a,
  \;
  v \in V^{(\ell)},\;
  a \in \mathbb{R}^{1 \times m_\ell},
  \label{eq:action}
\end{equation}
with $\rho^{(\ell)} \in \mathbb{R}^{d_\ell \times d_\ell}$ is the
Wigner–$D$ matrix \cite{thomas2018tensor} of order~$\ell$. One can realize that the group action $g$ only acts on
$V^{(\ell)}$, and the multiplicity space $\mathbb{R}^{1 \times m_\ell}$ is
invariant.

\paragraph{A Lightweight gating network}
In MMEA, we only take $\mathbf{h}^{(0)}$ as input  to maintain parameter efficiency. Then,  we design a two-layer MLP to obtain multiplicity-wise gain scalars. In particular, it consists of a bottleneck projection

\begin{equation}
  \mathbf{z}
  \;=\;
  \operatorname{SiLU}\!\bigl(
    W_{\!\downarrow}\,\mathbf{h}^{(0)}
    + \mathbf{b}_{\downarrow}
  \bigr),
  \qquad
  W_{\!\downarrow} \in \mathbb{R}^{r \times m_0},
  \label{eq:bottleneck}
\end{equation}
 and followed by an expansion:
\begin{equation}
  \bigl[
    \bm{\gamma}^{(0)},\,
    \bm{\gamma}^{(1)},\,
    \dots,\,
    \bm{\gamma}^{(L)}
  \bigr]
  =
  W_{\!\uparrow}\,\mathbf{z}
  + \mathbf{b}_{\uparrow}\,
  \label{eq:gains-update}
\end{equation}

\begin{equation}
  W_{\!\uparrow}
  \;\in\;
  \mathbb{R}^{\bigl(\sum_{\ell=0}^{L} m_\ell\bigr)\times r}\,,
  \label{eq:gains-shape}
\end{equation}

\noindent where each
$\bm{\gamma}^{(\ell)} \in \mathbb{R}^{m_\ell}$
assigns one scalar gain per multiplicity copy of order~$\ell$. $r$ is the bottleneck dimension used in the MLP. SiLU is Sigmoid Linear Unit \cite{elfwing2018sigmoid}, and $W_{\!\uparrow}$, $W_{\!\downarrow}$, $b_{\!\uparrow}$, $b_{\!\downarrow}$ are trainable parameters.

\paragraph{Equivariant modulation}
Write
$
  \mathbf{h}^{(\ell)}
  =
  \sum_{k=1}^{m_\ell}
  \mathbf{v}^{(\ell)}_{k} \otimes \mathbf{e}_k^{(\ell)}
$
with
$\mathbf{v}^{(\ell)}_{k} \in V^{(\ell)}$
and
$\{\mathbf{e}_k^{(\ell)}\}$ the standard basis of $\mathbb{R}^{m_\ell}$.
Define
\begin{align}
  \mathcal{A}_{\Gamma}^{(0)}
  \bigl(\mathbf{h}^{(0)}\bigr)
  &=
  \mathbf{h}^{(0)}
  +
  \bm{\gamma}^{(0)},
  \label{eq:scalar-update}
  \\[4pt]
  \mathcal{A}_{\Gamma}^{(\ell)}
  \bigl(\mathbf{h}^{(\ell)}\bigr)
  &=
  \sum_{k=1}^{m_\ell}
    \phi\!\bigl(\gamma^{(\ell)}_{k}\bigr)\,
    \mathbf{v}^{(\ell)}_{k}
    \otimes
    \mathbf{e}_k^{(\ell)},
  \qquad
  \ell \ge 1,
  \label{eq:tensor-update}
\end{align}
where
$\phi\!(x)=1+x$ (residual scaling)
or
$\phi\!(x)=e^{x}$ (positive scaling).
Then we obtain

\begin{equation}
  \mathcal{A}_{\Gamma}(\mathbf{h})
  :=
  \bigoplus_{\ell=0}^{L}
    \mathcal{A}_{\Gamma}^{(\ell)}\!\bigl(\mathbf{h}^{(\ell)}\bigr),
  \label{eq:adapter}
\end{equation}

\noindent yielding the modulated feature
$
  \mathbf{h}' := \mathcal{A}_{\Gamma}(\mathbf{h}) \in \mathcal{H}.
$

\subsection{The Equivariance Property of MMEA}
Now, we will show the equivariance property of our designed MMEA adapter. In other words, we will show that, for every $g \in \mathrm{SO}(3)$,
\begin{equation}
  \mathcal{A}_{\Gamma}\bigl(g \cdot \mathbf{h}\bigr)
  \;=\;
  g \cdot \mathcal{A}_{\Gamma}(\mathbf{h}).
  \label{eq:goal-equivariance}
\end{equation}

\paragraph{Step 1: The gains $\bm{\gamma}^{(\ell)}$ are invariant.}
Recall that the gains $\bm{\gamma}^{(\ell)}$ are produced solely from the scalar component $\mathbf{h}^{(0)} \in \mathbb{R}^{m_0}$ via the lightweight MLP:
\begin{equation}
  \mathbf{z}
  =
  \operatorname{SiLU}\!\bigl(
    W_{\!\downarrow}\,{\mathbf{h}^{(0)}}
    + \mathbf{b}_{\downarrow}
  \bigr),
\end{equation}
\begin{equation}
  \bigl[
    \bm{\gamma}^{(0)},\,
    \bm{\gamma}^{(1)},\,
    \dots,\,
    \bm{\gamma}^{(L)}
  \bigr]
  =
  W_{\!\uparrow}\,\mathbf{z}
  + \mathbf{b}_{\uparrow}.
\end{equation}
Since $\mathbf{h}^{(0)}$ transforms under the trivial irreducible representation ($\ell = 0$), we have
\begin{equation}
  g \cdot \mathbf{h}^{(0)} = \mathbf{h}^{(0)} \quad \forall\, g \in \mathrm{SO}(3).
\end{equation}
Consequently, $\mathbf{z}$ and hence every $\bm{\gamma}^{(\ell)}$ are unchanged by the group action:
\begin{equation}
  \bm{\gamma}^{(\ell)}(g \cdot \mathbf{h}^{(0)}) = \bm{\gamma}^{(\ell)}(\mathbf{h}^{(0)}), \qquad \forall\, \ell.
\end{equation}

\paragraph{Step 2: Each block $\mathcal{A}_{\Gamma}^{(\ell)}$ commutes with the group action.}
Decompose the $\ell$-th order feature as
\begin{equation}
  \mathbf{h}^{(\ell)}
  =
  \sum_{k=1}^{m_\ell}
  \mathbf{v}^{(\ell)}_{k} \otimes \mathbf{e}_k^{(\ell)},
  \qquad
  \mathbf{v}^{(\ell)}_{k} \in V^{(\ell)},
\end{equation}
the group action $g$ acts only on $V^{(\ell)}$:
\begin{equation}
  g \cdot \mathbf{h}^{(\ell)}
  =
  \sum_{k=1}^{m_\ell}
  \rho^{(\ell)}(g)\,\mathbf{v}^{(\ell)}_{k} \otimes \mathbf{e}_k^{(\ell)}.
\end{equation}
By definition, the adapter block is
\begin{align}
  \mathcal{A}_{\Gamma}^{(0)}(\mathbf{h}^{(0)})
  &= \mathbf{h}^{(0)} + \bm{\gamma}^{(0)}, \\[4pt]
  \mathcal{A}_{\Gamma}^{(\ell)}(\mathbf{h}^{(\ell)})
  &= \sum_{k=1}^{m_\ell}
     \phi\!\bigl(\gamma^{(\ell)}_k\bigr)\,
     \mathbf{v}^{(\ell)}_{k} \otimes \mathbf{e}_k^{(\ell)},
     \qquad \ell \ge 1.
\end{align}

For $\ell \ge 1$, because the gains $\gamma^{(\ell)}_k$ are group-invariant, applying $\mathcal{A}_{\Gamma}^{(\ell)}$ to the transformed features yields
\begin{align}
  \mathcal{A}_{\Gamma}^{(\ell)}\!\bigl(g \cdot \mathbf{h}^{(\ell)}\bigr)
  &= \sum_{k=1}^{m_\ell}
     \phi\!\bigl(\gamma^{(\ell)}_k\bigr)\,
     \bigl(\rho^{(\ell)}(g)\,\mathbf{v}^{(\ell)}_{k}\bigr) \otimes \mathbf{e}_k^{(\ell)} \\
  &= \rho^{(\ell)}(g)
     \Biggl(
       \sum_{k=1}^{m_\ell}
       \phi\!\bigl(\gamma^{(\ell)}_k\bigr)\,
       \mathbf{v}^{(\ell)}_{k} \otimes \mathbf{e}_k^{(\ell)}
     \Biggr) \\
  &= g \cdot \mathcal{A}_{\Gamma}^{(\ell)}\!\bigl(\mathbf{h}^{(\ell)}\bigr).
\end{align}

For $\ell=0$, according to Step 1, it is easy to check 

\begin{align}
  \mathcal{A}_{\Gamma}^{(0)}(g \cdot \mathbf{h}^{(0)})
   = g \cdot \mathcal{A}_{\Gamma}^{(0)}(\mathbf{h}^{(0)}).
\end{align}

Therefore, 
\begin{align}
  \mathcal{A}_{\Gamma}\bigl(g \cdot \mathbf{h}\bigr)
  &= \bigoplus_{\ell=0}^{L}
     \mathcal{A}_{\Gamma}^{(\ell)}\!\bigl(g \cdot \mathbf{h}^{(\ell)}\bigr) \notag \\
  &= \bigoplus_{\ell=0}^{L}
     g \cdot \mathcal{A}_{\Gamma}^{(\ell)}\!\bigl(\mathbf{h}^{(\ell)}\bigr) \notag \\
  &= g \cdot \mathcal{A}_{\Gamma}(\mathbf{h}).
\end{align}

establishing \eqref{eq:goal-equivariance}. \qed

\section{Experiments}

In this section, we first introduce the experimental setup. Next, we evaluate our method on several popular datasets and compare it against baseline approaches. Finally, we conduct ablation study and assess data efficiency. Additional experiments, including training efficiency, computational overhead and deviation analysis are provided in the Appendix.

\subsection{Experimental Setup}

\subsubsection{Pretrained model}  

Same as previous work, we adopt the MACE \cite{Batatia2022mace} as the backbone to evaluate the effectiveness of our proposed MMEA.
The pretrained model employed in this work is MACE-OFF \cite{kovacs2025mace}. 
The corresponding hyperparameters are summarized in Table~\ref{tab:mace_hyp}.
More details about MACE are provided in the supplementary materials. 

\begin{table}[hb]
\centering
\begin{tabular}{lc}
\toprule
Hyperparameter & Value \\
\midrule
Correlation order                       & 3   \\
Cutoff radius      & 5.0 \\
Max order for input      & 3   \\
Max order for product layer      & 1   \\
Number of hidden channels               & 128 \\
Number of interaction layers            & 2   \\
Number of radial basis functions        & 8   \\
\bottomrule
\end{tabular}
\caption{Hyperparameter settings of model.}
\label{tab:mace_hyp}
\end{table}

\subsubsection{Datasets} 

Following the experimental setup of MACE, we selected benchmark datasets from the molecular potential prediction domain. More detailed information about these datasets is provided in the supplementary materials. The target properties for each dataset are energy (E) and forces (F). These datasets include:

Firstly, the revised MD-17 dataset \cite{christensen2020role}. It consists of ten small organic molecules, for which 100,000 structures were computed at DFT accuracy using a very tight SCF convergence and very dense DFT integration grid. The structures were recomputed from the original MD-17 dataset.

Secondly, the 3BPA dataset \cite{kovacs2021linear}. It contains DFT train test splits of a flexible drug-like organic molecule sampled from different temperature molecular dynamics trajectories. The models is trained on 500 snapshots sampled at 300K and tested on three independent test sets for each temperature (300K, 600K, 1200K). The models can also be tested on the challenging task of computing the energy along dihedral rotations of the molecule. 
Thirdly, the acetylacetone (AcAc) dataset \cite{Batatia2022mace}. It contains trajectories of a small reactive molecule sampled at different temperature. The task is to train on snapshot sampled at 300K and test on independent test sets sampled at 300K and 600K. Moreover the extrapolation is measured both in temperature and along two internal coordinates of the molecule, the hydrogen transfer path and a partially conjugated double bond rotation, which has a very high barrier for rotation.

\subsubsection{Experimental Method}

We compare the Mean Absolute Error (MAE) / Root Mean Square Error (RMSE) of energy  and forces for pretrained models with (i) full-parameter fine-tuning (Full), (ii) ELoRA fine-tuning, and (iii) MMEA fine-tuning (our method). Except for the rMD17 dataset, we also recorded the performance of (iiii) trained from scratch (Scratch).
All fine-tuning methods in each experiment employ the same pretrained model, dataset, and training hyperparameters. Overall, our method follows the experimental setup of ELoRA. The key distinction is that all equivariant linear layers previously fine-tuned with ELoRA are now fine-tuned using our proposed MMEA method. Tensor Product layers and a small number of scalar networks are consistent with the original ELoRA configuration.
Unless otherwise specified, the low‑rank components in both methods are set to rank $r=16$.
Each experiment is conducted over three runs.

For the rMD17 dataset, since the performance of few-shot learning models is significantly influenced by data splits, we randomly sampled 50 configurations from the officially provided training set to construct our training subset, ensuring that all methods use the exact same data, thereby enabling fair training. For the remaining datasets, the results for full-parameter and ELoRA fine-tuning are directly taken from previous work \cite{wang2025elora}.

The experiments were carried out with the PyTorch framework \cite{paszke2019pytorch} on a server running Ubuntu 22.04.5 LTS. We utilized a single NVIDIA A100 GPU for training and evaluation.
More training details are provided in the supplementary materials.

\begin{table}[t]
  \centering

  \sisetup{detect-weight=true,detect-inline-weight=math}
  \small{
  \begin{tabular}{>{\raggedright\arraybackslash}p{1.5cm}
                cccc}
    \toprule
    \multicolumn{2}{c}{rMD17} & \multicolumn{3}{c}{MAE$\downarrow$} \\
    \cmidrule(lr){3-5}
    Molecule & Metric & {Full} & {ELoRA} & {MMEA } \\
    \midrule
   \multirow{2}{*}{Aspirin}       
        & E &  $9.7$  &  $8.0\,(\downarrow18\%)$ & $\mathbf{7.3}\,(\downarrow25\%)$ \\
        & F & $23.9$  & $18.3\,(\downarrow23\%)$ & $\mathbf{16.4}\,(\downarrow31\%)$ \\ \addlinespace
    \multirow{2}{*}{Azobenzene}    
        & E &  $4.6$  &  $4.0\,(\downarrow13\%)$ & $\mathbf{3.9}\,(\downarrow15\%)$ \\
        & F & $14.8$  & $12.6\,(\downarrow15\%)$ & $\mathbf{11.9}\,(\downarrow20\%)$ \\ \addlinespace
    \multirow{2}{*}{Benzene}       
        & E &  $0.3$  & $\mathbf{0.2}\,(\downarrow33\%)$ & $\mathbf{0.2}\,(\downarrow33\%)$ \\
        & F &  $2.4$  &  $1.6\,(\downarrow33\%)$ & $\mathbf{1.4}\,(\downarrow42\%)$ \\ \addlinespace
    \multirow{2}{*}{Ethanol}       
        & E &  $2.9$  & $\mathbf{2.3}\,(\downarrow21\%)$ & $\mathbf{2.3}\,(\downarrow21\%)$ \\
        & F & $14.4$  & $11.5\,(\downarrow20\%)$ & $\mathbf{10.8}\,(\downarrow25\%)$ \\ \addlinespace
    \multirow{2}{*}{Malonaldehyde} 
        & E &  $6.8$  &  $6.6\,(\downarrow3\%)$  & $\mathbf{6.4}\,(\downarrow6\%)$ \\
        & F & $25.4$  & $22.3\,(\downarrow12\%)$ & $\mathbf{21.6}\,(\downarrow15\%)$ \\ \addlinespace
    \multirow{2}{*}{Naphthalene}   
        & E &  $1.8$  &  $1.4\,(\downarrow22\%)$ & $\mathbf{1.2}\,(\downarrow33\%)$ \\
        & F &  $8.1$  &  $6.0\,(\downarrow26\%)$ & $\mathbf{5.7}\,(\downarrow30\%)$ \\ \addlinespace
    \multirow{2}{*}{Paracetamol}   
        & E &  $6.5$  &  $4.7\,(\downarrow28\%)$ & $\mathbf{4.3}\,(\downarrow34\%)$ \\
        & F & $20.3$  & $14.5\,(\downarrow29\%)$ & $\mathbf{13.3}\,(\downarrow34\%)$ \\ \addlinespace
    \multirow{2}{*}{Salicylic}     
        & E &  $4.3$  &  $3.2\,(\downarrow26\%)$ & $\mathbf{2.9}\,(\downarrow33\%)$ \\
        & F & $17.2$  & $13.3\,(\downarrow23\%)$ & $\mathbf{12.0}\,(\downarrow30\%)$ \\ \addlinespace
    \multirow{2}{*}{Toluene}       
        & E &  $1.8$  &  $1.4\,(\downarrow22\%)$ & $\mathbf{1.2}\,(\downarrow33\%)$ \\
        & F &  $8.8$  &  $6.2\,(\downarrow30\%)$ & $\mathbf{5.4}\,(\downarrow39\%)$ \\ \addlinespace
    \multirow{2}{*}{Uracil}        
        & E &  $2.9$  &  $2.1\,(\downarrow28\%)$ & $\mathbf{2.0}\,(\downarrow31\%)$ \\
        & F & $15.8$  & $12.3\,(\downarrow22\%)$ & $\mathbf{10.7}\,(\downarrow32\%)$ \\ \addlinespace
    \bottomrule
  \end{tabular}
  }
  \caption{Experimental results on 10 Molecules of rMD17. Each model is trained on 50 samples. Results are rounded to one decimal place, the decrease ratio is calculated based on the results of full fine-tuning, and  the best results are in bold.}
    \label{tab:md_results}
\end{table}

\subsection{Experimental results}

\subsubsection{Results on rMD17}
The rMD17 dataset here serves as an effective benchmark for evaluating model performance under few‑shot learning scenarios. The results (Table~\ref{tab:md_results}) show that our proposed MMEA method consistently outperforms previous baselines. Compared to ELoRA, MMEA achieves an additional average improvement of approximately 6.6\% on the energy metric and 8.7\% on the force metric.
 Overall, MMEA compresses the average energy and force MAEs by 6–8\%, achieves 10–25\% improvements on high‑error molecules, and establishes new state‑of‑the‑art results. These findings demonstrate that MMEA can more effectively leverage a model’s existing knowledge to attain high‑precision predictions in extremely low‑sample scenarios.

\subsubsection{Results on 3BPA}
The results of evaluation on 3BPA Dataset is shown in Table \ref{tab:3bpa_results}. The results show that MMEA yields a modest improvement over ELoRA on the 300~K test, that has the same distribution as the training set. When the same model is evaluated directly at higher temperatures, it consistently maintains lower energy and force errors, which demonstrating MMEA’s ability to generalize and mitigate forgetting of pretrained knowledge. Furthermore, on the dihedral slice test, MMEA reduces both energy  and force error significantly, indicating a  more accurate reconstruction of the potential energy surface and improved stability and precision in conformer predictions.

\begin{table*}[htbp]
\centering
\begin{tabular}{cccccc}
\toprule
\multicolumn{2}{c}{3BPA} & \multicolumn{4}{c}{RMSE$\downarrow$} \\
\cmidrule(lr){3-6}
Condition & Metric & Scratch & Full & ELoRA & MMEA  \\
\midrule
\multirow{2}{*}{300~K}  & E & $3.0\pm0.2$    & $3.3\pm0.03$    & $3.0\pm0.05$      & $\mathbf{2.7}\pm0.03$     \\
                        & F & $8.8\pm0.3$    & $7.8\pm0.01$    & $\mathbf{7.5}\pm0.05$  & $\mathbf{7.5}\pm0.01$     \\ \addlinespace
\multirow{2}{*}{600~K}  & E & $9.7\pm0.5$    & $7.3\pm0.04$    & {$\mathbf{6.5}\pm0.10$}  & $\mathbf{6.5}\pm0.03$     \\ 
                        & F & $21.8\pm0.6$   & $16.6\pm0.05$   & $15.5\pm0.12$     & $\mathbf{15.4}\pm0.14$    \\ \addlinespace
\multirow{2}{*}{1200~K} & E & $29.8\pm1.0$   & $20.3\pm0.17$   & $17.6\pm0.11$     & $\mathbf{17.1}\pm0.14$    \\
                        & F & $62.0\pm0.7$   & $48.7\pm0.56$   & $42.0\pm0.51$     & $\mathbf{39.7}\pm0.12$    \\ \addlinespace
\multirow{2}{*}{Dihedral Slices} & E & $7.8\pm0.6$ & $7.3\pm0.28$    & $5.9\pm0.28$      & $\mathbf{5.6}\pm0.20$     \\
                        & F & $16.5\pm1.7$   & $12.3\pm0.10$   & $11.4\pm0.17$     & $\mathbf{10.6}\pm0.17$    \\ 
\bottomrule
\end{tabular}
\caption{Experimental results under three temperature conditions for 3BPA. The model was trained at 300~K and tested at 300~K, 600~K, and 1200~K to evaluate its generalization performance. Standard deviations are computed over three runs. The best results are in bold.}
\label{tab:3bpa_results}
\end{table*}

\subsubsection{Results on AcAc}
The results of evaluation on AcAc Dataset is shown in Table \ref{tab:acac_results}.
At 300~K, MMEA  achieves the lowest energy and force RMSE, slightly outperforming other baselines. When generalizing to 600~K, it continues to maintain low errors, This demonstrates that it better leverages the pre‑trained model’s knowledge to achieve superior generalization performance, further demonstrating the effectiveness. We also evaluated  MMEA  with rank set to 32 on this dataset. By increasing its parameter count, we observed a slight performance improvement.

The experiments across multiple datasets reveal the effectiveness of our method. On the rMD17 benchmark, we demonstrate its few‑shot learning capability under limited data. On the 3BPA and AcAc benchmarks, we demonstrate its superior generalization performance. This outcome is intuitive: by preserving the multiplicity basis, our method more fully exploits the knowledge in the pretrained model.


\begin{table*}[htbp]
\centering

\begin{tabular}{ccccccc}
\toprule
\multicolumn{2}{c}{AcAc} & \multicolumn{5}{c}{RMSE$\downarrow$} \\
\cmidrule(lr){3-7}
Condition & Metric & Scratch & Full & ELoRA & MMEA($r=16$) & MMEA($r=32$) \\
\midrule
\multirow{2}{*}{300~K}  & E & $0.9\pm0.03$   &  $1.0\pm0.02$  &  ${0.8\pm0.03}$  & $\mathbf{0.7}\pm0.04$ & $\mathbf{0.7}\pm0.02$ \\
                        & F & $5.1\pm0.10$   &  $5.1\pm0.07$  &  $4.5\pm0.06$  & $\underline{4.4\pm0.02}$ & $\mathbf{4.2}\pm0.02$ \\ \addlinespace
\multirow{2}{*}{600~K}  & E & $4.6\pm0.3$    &  $5.8\pm0.28$  &  $3.9\pm0.33$  & $\underline{3.6\pm0.06}$ & $\mathbf{3.2}\pm0.10$\\ 
                        & F & $22.4\pm0.9$   & $16.4\pm0.70$  & $13.6\pm0.26$  & $\underline{13.2\pm0.11}$ & $\mathbf{13.0}\pm0.23$\\ 
\bottomrule

\end{tabular}
\caption{Experimental results under two temperature conditions for AcAc. The model was trained at 300~K and tested at 300~K and 600~K to evaluate its generalization performance. Standard deviations are computed over three runs. The best results are in {bold} and the second best are {underlined}.}
\label{tab:acac_results}
\end{table*}

\subsection{Parameter Efficiency}

We compare the parameter efficiency of Full fine-tuning, ELoRA, and our proposed MMEA with two different rank settings. The result is shown in Table~\ref{tab:param_budget}. Our method ($r=16$) updates approximately 20\% of the parameters used in full fine-tuning, corresponding to about 85\% of ELoRA's parameter budget, while achieving superior fine-tuning performance. 
When $r$ is set to 32, the parameter count of our method  exceeds ELoRA, but as shown in Table \ref{tab:acac_results}, it can further improve performance.

\begin{table}[htbp]
\centering
\begin{tabular}{lcrr}
\toprule
{Method} & $r$&{\# params} & {\% of Full FT} \\
\midrule
Full & /&751896 & 100.0\% \\
ELoRA &16& 175880 & 23.4\%  \\
{MMEA } &16& 151354 & 20.1\%  \\
{MMEA } &32&  201258 & 26.7\%  \\
\bottomrule
\end{tabular}
\caption{Trainable parameter for fine-tuning on rMD17.}
\label{tab:param_budget}
\end{table}

\subsection{Ablation Study}

Table~\ref{tab:ablation} presents the ablation study conducted on the rMD17-Aspirin dataset. 
We removed several key components individually to examine their effects on model performance. w/o nonlinear activation means removing the activation function; w/o input-head reuse indicates disabling the input-head reuse mechanism, where scalar outputs ($l=0$) and higher-order tensor outputs ($l>0$) are processed using separate MLPs; w/o high-order refers to omitting the high-order adjustments, i.e., tuning only the scalar information; w/o scalar modulation means no modulation is applied to the scalar channels, which significantly degrades performance; and use same modulation applies identical modulation to all higher-order channels. These results collectively demonstrate that each proposed component plays an important role in improving model performance.

Additionally, we compared two other common fine‑tuning strategies. The first (Readout) tunes only the readout (i.e., the layer directly producing the final outputs), updating just 0.3\% of the model’s parameters. While extremely parameter‑efficient, this approach greatly limits the model’s learning capacity and yields poor fine‑tuning performance. The second (Adapter) employs a conventional bottleneck for fine‑tuning; by mixing features from different irreducible representations, it breaks the model’s equivariance. As a result, the model cannot fully leverage its pre‑trained knowledge and again suffers degraded performance. 
Both of these methods underperform full fine‑tuning, ELoRA, and MMEA.

\begin{table}[htbp]
  \centering
  \begin{tabular}{lcc}
    \toprule
    \multicolumn{1}{c}{rMD17-Aspirin} & \multicolumn{2}{c}{MAE$\downarrow$} \\
    \cmidrule(lr){2-3}
    Method & Energy & Forces \\
    \midrule
    MMEA                          & $\mathbf{7.3}$                     & $\mathbf{16.4}$                   \\
    Full                          & $9.7$                     & $23.9$                   \\
\midrule
    w/o nonlinear activation      & $7.6$    & $16.4$  \\
    w/o input-head reuse          & $9.2$   & $16.7$  \\
    w/o scalar modulation             & $12.9$  & $30.5$  \\
    w/o high-order modulation     & $8.3$   & $16.6 $  \\
    shared high-order modulation            & $7.6$   & $16.6$  \\
    \midrule
    Readout  &$ 23.8$ & $ 36.8$\\
    Adapter &  $11.0$&    $26.3$\\
\bottomrule
  \end{tabular}
  \caption{Ablation study results on rMD17-Aspirin Dataset.}
  \label{tab:ablation}
\end{table}

\begin{figure}[htbp]
    \centering
    \includegraphics[width=0.9\linewidth,trim={0 20 0 20},clip]{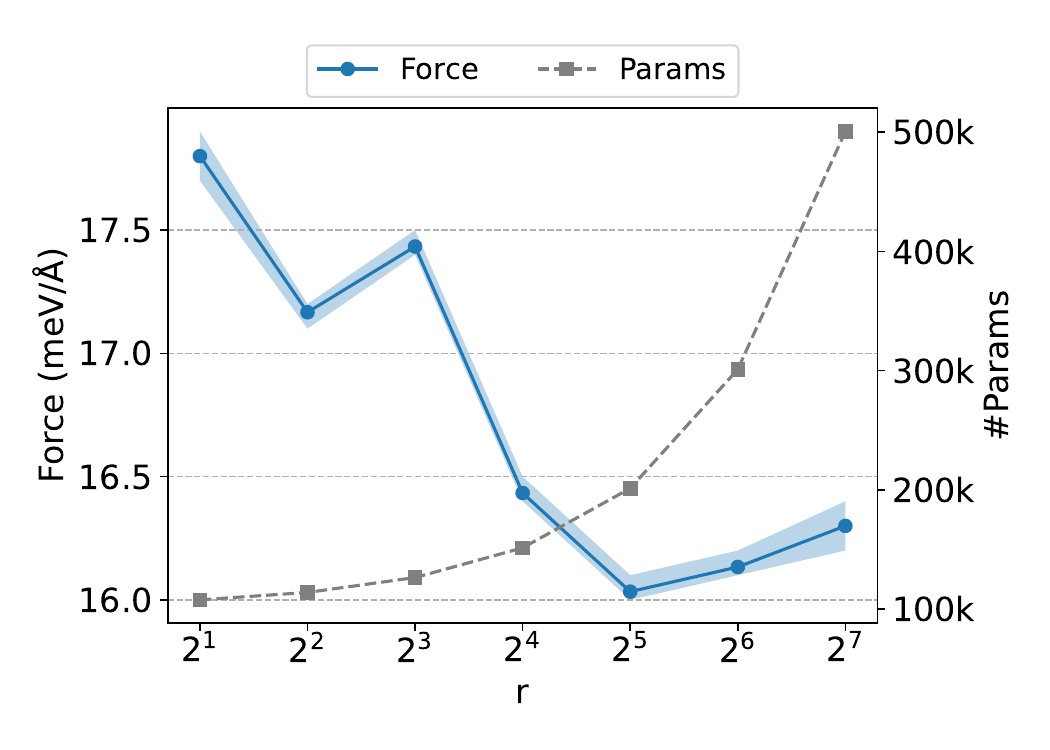} 
    \caption{Force MAE and parameter counts across different ranks. Results are reported on the rMD17‑Aspirin dataset.}
    \label{fig:rank}
\end{figure}

Like many PEFT approaches, the number of trainable parameters in MMEA is governed by the rank of its bottleneck structure. To examine how this hyperparameter influences model performance, we evaluate forces prediction on the rMD17-Aspirin dataset. As illustrated in Fig.~\ref{fig:rank}, increasing the rank initially improves accuracy, but beyond a certain point leads to performance degradation. This trend reflects a common trade-off: low-rank configurations may lack expressiveness and underfit, whereas higher ranks risk overfitting due to excess capacity. To enable direct comparison with prior work, we use $r=16$ in our main experiments, to achieve stronger results with fewer parameters. Nevertheless, we observe that a larger setting, $r=32$, yields better accuracy in this task, suggesting that further improvements are possible when the parameter budget allows. Therefore, the optimal rank should be treated as a task‑specific hyperparameter and tuned accordingly for each downstream application.

\section{Conclusion}
Equivariant pretrained models are currently indispensable for machine-learning interatomic potentials as they preserve the underlying physical symmetries needed for accurate, reliable atomistic simulations. Fine-tuning these models improves downstream accuracy, especially in low-data scenarios, but standard PEFT techniques break equivariance and thus underperform. Approaches such as ELoRA address this by explicitly retaining symmetry during adaptation, yet their parameter efficiency and performance can still be improved.

In a well-trained equivariant GNN, every irreducible order and its associated multiplicity channels already form a robust representation; effective adaptation therefore requires only subtle, per-channel scaling. Leveraging this insight, we introduce MMEA, a PEFT method that uses scalar gating to modulate each order-multiplicity channel independently, avoiding any mixing within an irreducible order. This preserves equivariance, exploits the pretrained knowledge more directly, and attains stronger performance with even fewer trainable parameters. MMEA delivers state-of-the-art results on the rMD17, 3BPA, and AcAc benchmarks, demonstrating its effectiveness and efficiency.

\paragraph{Limitations}
Although MMEA method demonstrates superior performance on various molecular datasets, we observed certain limitations under specific conditions. Firstly, when the target system significantly deviates from the pretrained model distribution, our method underperforms compared to ELoRA. This occurs primarily because ELoRA permits the mixing of multiplicities, offering greater flexibility and making it easier to adaptively fit challenging datasets. However, for entirely novel systems with minimal overlap to the pretrained dataset, full-parameter fine-tuning remains the most effective strategy, as shown in Table~\ref{tab:ood}. Therefore, and quite intuitively, when the distribution gap between the pre-training model and the downstream task is large, one should train from scratch or perform full fine-tuning; when the gap is moderate or small, leveraging ELoRA or MMEA can yield better performance.
Secondly, unlike ELoRA, whose learned weights can be seamlessly merged back into the backbone model parameters, our approach does not allow parameter merging. Consequently, our method introduces slight additional computational overhead during inference.

\begin{table}[ht]
\centering

\begin{tabular}{>{\raggedright\arraybackslash}p{1.5cm}
                c
                  c
                  c
                  c}
    \toprule
    \multicolumn{2}{c}{rMD17} & \multicolumn{3}{c}{MAE$\downarrow$} \\
    \cmidrule(lr){3-5}
    Molecule & Metric & {Full} & {ELoRA} & {MMEA } \\
\midrule
\multirow{2}{*}{Aspirin}     & E & $\mathbf{11.4}$      & {$12.5$}                 & {$13.3$}      \\
                             & F & $\mathbf{29.4}$      & {$31.1$}     & $31.9$                  \\
\addlinespace

\multirow{2}{*}{Uracil}      & E & $\mathbf{3.4}$       & {$3.5$}                  & {$3.6$}       \\
                             & F & $\mathbf{19.7}$      & {$20.7$}     & $21.4$                  \\
\bottomrule

\end{tabular}
\caption{Experimental results using MACE-MP \cite{batatia2023foundation} pretrained on inorganic compounds to predict organic compounds, assessing extreme out‑of‑distribution  performance. The best results are in bold.}
\label{tab:ood}
\end{table}

\paragraph{Future Work}
Future research directions include the following aspects:
Firstly, developing efficient equivariant methods for Fully Connected Tensor Product. Since the tensor product involves multiple inputs, how to implement gating effectively remains an open question. We have attempted to apply gating based on the output, but this approach has not shown significant benefits.
Secondly, designing a generalizable fine-tuning method. A promising direction is to combine the strengths of the three existing approaches to create a unified method that consistently demonstrates strong performance on data distributions that differ from the pretraining distribution.
Thirdly, improving fine-tuning efficiency. Currently, fine-tuning convergence remains relatively slow. Future work could explore the development of more efficient fine-tuning schemes that enable faster convergence.

\section*{Acknowledgments}
This research was supported by the Research Grants Council of Hong Kong (Grant No. T42-513/24-R).

\bibliography{aaai2026}

\input{appendix}

\end{document}

%% file: appendix.tex
\appendix

\section{Experimental Details}

\subsection{Pretrain Data}
MACE-OFF was pretrained on SPICE \cite{eastman2023spice}. It contains small molecules of up to 50 atoms and involves ten organic  chemical elements: H, C, N, O, F, P, S, Cl, Br, and I. To facilitate the learning of intramolecular non-bonded interactions, it also uses a few larger molecules of 50-90 atoms randomly selected from the QMugs dataset \cite{isert2022qmugs}.

\subsection{Evaluation Datasets}

Here, we present additional details regarding the datasets used in the experiment.

The revised MD17 (rMD17) dataset \cite{christensen2020role}  is constructed based on the original MD17 dataset.
100K structures were randomly chosen for each type of molecule present in the MD17 dataset. Subsequently, the
single-point force and energy calculations were performed for each of these structures using the PBE/def2-SVP level of theory. The calculations were conducted with tight SCF convergence and a dense DFT integration grid, significantly minimizing noise.

The 3BPA dataset \cite{kovacs2021linear} is a dataset consisting of a flexible druglike molecule 3-(benzyloxy)pyridin-2-amine. This dataset features complex dihedral potential energy surface with many local minima, which can be challenging to approximate using classical or ML force fields. The configuration were sampled from short (0.5 ps) MD simulations using the ANI-1x force field to perturb the toward lower potential energies. Furthermore, long 25 ps MD simulation were performed at three different temperatures (300, 600, and 1200 K) using the Langevin thermostat and a 1 fs time step. The final configurations were re-evaluated using ORCA at the DFT level of theory using the $\omega$B97X exchange correlation functional and the 6-31G(d) basis set.

The acetylacetone (AcAc) dataset \cite{batzner20223} focuses on the potential energy surface modeling of the acetylacetone molecule, which presents significant challenges due to its complex internal coordinate transitions. The training set consists of 500 configurations sampled every 1~ps from a long molecular dynamics simulation performed at 300~K using a Langevin thermostat at the semi-empirical GFN2-xTB level of theory. These configurations were re-evaluated using DFT with the PBE exchange-correlation functional, D3 dispersion correction, the def2-SVP basis set, and VeryTightSCF convergence settings, implemented in the {ORCA} electronic structure package.
The test set is designed to evaluate model extrapolation both in temperature (from 300~K to 600~K) and along internal molecular coordinates. In particular, it probes two high-barrier transition modes: intramolecular hydrogen transfer and rotation around a partially conjugated double bond. Due to the inclusion of these complex dynamical features, the AcAc dataset provides a stringent benchmark for assessing the generalization capabilities of machine learning potentials. 

\subsection{Hyperparameter settings}

The experimental hyperparameters are listed in Table~\ref{tab:exp_hyp}. These settings are consistent with those used in the MACE and ELoRA experiments.

\begin{table}[htbp]
\centering
\caption{Hyperparameter settings for MACE.}
\begin{tabular}{lc}
\toprule
Hyperparameter                 & Value             \\
\midrule
$E_0$                           & average          \\
loss                          & ef (weighted)              \\
learning rate          & 0.005           \\
forces weight                  & 1000             \\
energy weight                  & 1                 \\
weight decay                   & 1e-8             \\
gradient clipping & 100               \\
batch size                     & 5              \\
max epochs                     & 500          \\
scheduler patience             & 5                 \\
EMA decay                      & 0.995            \\
\bottomrule
\end{tabular}
\label{tab:exp_hyp}
\end{table}

\section{Addtional Experiments}

\subsection{Training Efficiency}

\begin{figure}[htbp]
  \centering
  \includegraphics[width=0.9\linewidth]{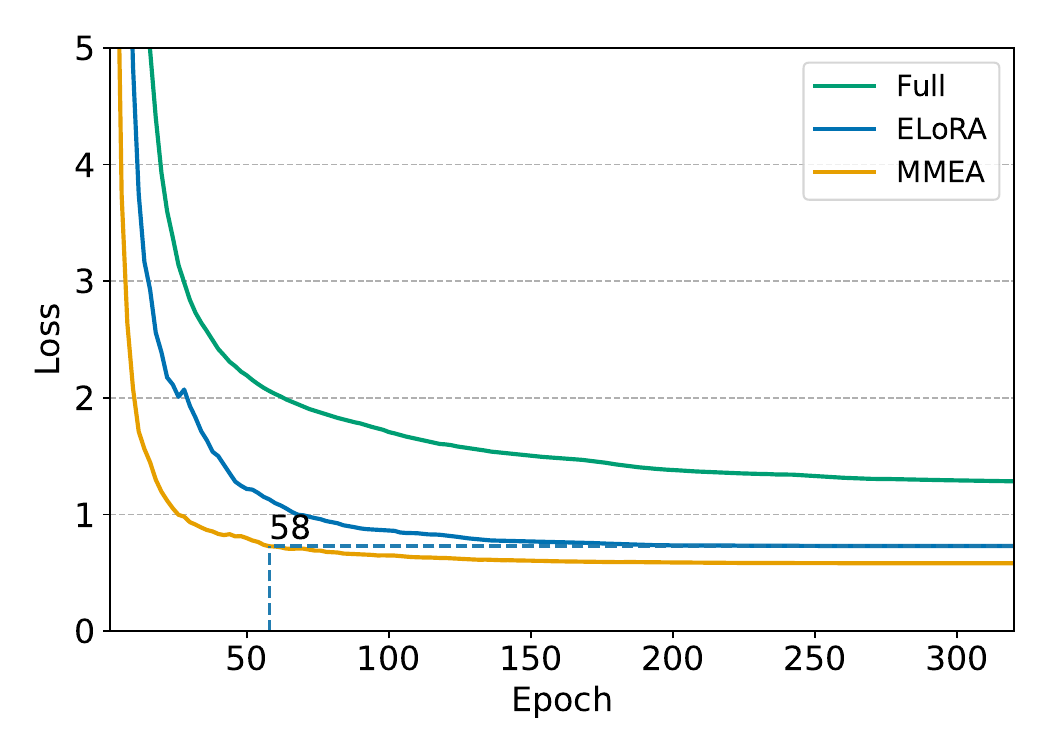}
  \caption{Validation loss convergence for ELoRA versus MMEA  on rMD17-Aspirin Dataset. A dashed line marks the point at epoch 58 where MMEA reaches ELoRA’s near final loss at around epoch 200.}
  \label{fig:convergence_comparison}
\end{figure}

We tested the model’s convergence speed on the Aspirin dataset. Our method exhibits significantly faster convergence. As shown in Fig.~\ref{fig:convergence_comparison}, representing the improvement in training efficiency.

\subsection{Computational Overhead of Inference}

The additional computations introduced by MMEA during the forward pass include a lightweight gated MLP based on the $\ell = 0$ scalar channel, which operates only on the scalar channels. Its complexity scales linearly with the rank and the number of scalar channels.  
Furthermore, MMEA performs element-wise channel-wise multiplications  for each irreducible representation channel.  
Overall, the computational complexity consists of two parts: 
$O\!\left(r \sum_{\ell=0}^{L} m_\ell\right)$
and $
O\!\left(\sum_{\ell=0}^{L} m_\ell (2\ell + 1)\right)$, $L$ for the max order.

Since the dominant computational cost of equivariant GNNs arises from tensor products and message passing, these two additional operations incur only a marginal overhead compared to the backbone cost.

To assess the inference efficiency, we performed an experiment measuring the inference latency, as shown in Table \ref{tab:coi}. Compared to ELoRA-Merged, the inference latency of MMEA increases by approximately 2.1\%. A detailed theoretical discussion will be provided accordingly.

\begin{table}[htbp]
\centering
\begin{tabular}{lcc}
\toprule
{Model} & Time/Sample (s) & {Throughput (samples/s)} \\
\midrule
{ELoRA} & 0.0626 & 15.99 \\
{MMEA} & 0.0639 & 15.66 \\
\bottomrule
\end{tabular}
\caption{Inference latency comparison between ELoRA and MMEA.}
\label{tab:coi}
\end{table}

\subsection{Deviation Analysis}
To better understand how the adaptations of ELoRA and MMEA affect the underlying equivariant representations, we conduct a numerical study of the angular deviation between the adapted models and the pretrained backbone at the interaction layers. For a pair of non-zero vectors $x$ and $y$ from the pretrained and adapted models, respectively, we define
$$
\theta(x, y)
= \arccos\left(
\left\langle
\frac{x}{\lVert x \rVert_2},
\frac{y}{\lVert y \rVert_2}
\right\rangle
\right),
$$
and report the mean and median of $\theta$ (in degrees) over all samples and channels.

As summarized in Table~\ref{tab:angular_deviation}, ELoRA consistently exhibits larger angular deviations than MMEA at both interaction layers, in terms of both mean and median. This indicates that ELoRA perturbs the pretrained equivariant representations more aggressively, which aligns with its design of mixing multiplicities within each order to increase flexibility at the cost of a higher risk of overfitting. In contrast, MMEA employs per-channel scalar gating, resulting in smaller deviations while still improving performance, suggesting a more controlled use of degrees of freedom.

\begin{table}[htbp]
\centering
\begin{tabular}{lcccc}
\toprule
Layer & Method & Mean ($^\circ$) & Median ($^\circ$) \\
\midrule
Interaction 1 & ELoRA & 0.3303 & 0.2068 \\
              & MMEA  & 0.2631 & 0.1642 \\
\addlinespace
Interaction 2 & ELoRA & 0.3425 & 0.2383 \\
              & MMEA  & 0.2364 & 0.1649 \\
\bottomrule
\end{tabular}
\caption{Mean and median angular deviation (in degrees) between the adapted models and the pretrained backbone at the interaction layers.}
\label{tab:angular_deviation}
\end{table}

\section{Significance Analysis}

We conducted paired two-sided t-tests on the MAE of energy (E) and force (F) predictions for 10 molecules in the rMD17 dataset to assess (1) the performance improvements of ELoRA and MMEA relative to full fine-tuning, and (2) the difference between ELoRA and MMEA. The results are summarized in Table~\ref{tab:significance}. All p-values are well below 0.05, indicating that both adapter methods yield highly significant improvements over full fine-tuning and that MMEA outperforms ELoRA in both metrics.

\begin{table}[htbp]
\centering

\begin{tabular}{lccr}
\toprule
Comparison                & Metric     & t‑statistic & p‑value \\
\midrule
ELoRA vs Full             & E          &  4.122      & 0.0026  \\
ELoRA vs Full             & F          &  6.649      & 0.0001  \\ \addlinespace
MMEA vs Full              & E          &  4.076      & 0.0028  \\
MMEA vs Full              & F          &  6.546      & 0.0001  \\ \addlinespace
ELoRA vs MMEA             & E          &  3.317      & 0.0090  \\
ELoRA vs MMEA             & F          &  5.423      & 0.0004  \\
\bottomrule
\end{tabular}
\caption{Significance analysis results among three finetuning methods.}
\label{tab:significance}
\end{table}

\section{Backbone model}

\subsection{Architecture}

\label{app:mace}

The MACE architecture has rapidly gained traction in the atomistic machine-learning community. The MACE model follows the general framework of MPNNs. The key innovation is a new message construction mechanism: It expands the messages $m_i^{(t)}$ in a hierarchical body order expansion,

\begin{align}
m_i^{(t)} &= \sum_j u_1 \left( \sigma_i^{(t)} ; \sigma_j^{(t)} \right) + \sum_{j_1,j_2} u_2 \left( \sigma_i^{(t)} ; \sigma_{j_1}^{(t)}, \sigma_{j_2}^{(t)} \right) \nonumber \\
&\quad + \cdots + \sum_{j_1,\ldots,j_\nu} u_\nu \left( \sigma_i^{(t)} ; \sigma_{j_1}^{(t)}, \ldots, \sigma_{j_\nu}^{(t)} \right),
\end{align}

where the $u$ functions are learnable, the sums run over the neighbors of $i$, and $\nu$ is a hyper-parameter corresponding to the maximum correlation order, the body order minus 1, of the message function with respect to the states. 

\paragraph{Message Construction}
At each iteration, MACE first embed the edges using a learnable radial basis $R_{k l_1 l_2 l_3}^{(t)}$, a set of spherical harmonics $Y_{l_1}^{m_1}$, and a learnable embedding of the previous node features $h_{j,k l_2 m_2}^{(t)}$ using weights $W_{k k' l_2}^{(t)}$. The $A_i^{(t)}$-features are obtained by pooling over the neighbours $\mathcal{N}(i)$ to obtain permutation invariant 2-body features whilst, crucially, retaining full directional information, and thus, full information about the atomic environment:

\begin{equation}
\begin{split}
A_{i,k l_3 m_3}^{(t)} = \sum_{l_1 m_1, l_2 m_2} C_{l_1 m_1, l_2 m_2}^{l_3 m_3} \\
\cdot \sum_{j \in \mathcal{N}(i)} R_{k l_1 l_2 l_3}^{(t)}(r_{ji}) Y_{l_1}^{m_1}(\hat{r}_{ji}) \\
\cdot \sum_{k'} W_{k k' l_2}^{(t)} h_{j,k' l_2 m_2}^{(t)},
\end{split}
\end{equation}

where $C_{l_1 m_1, l_2 m_2}^{l_3 m_3}$ are the standard Clebsch-Gordan coefficients ensuring that $A_{i,k l_3 m_3}^{(t)}$ maintain the correct equivariance, $r_{ji}$ is the (scalar) interatomic distance, and $\hat{r}_{ji}$ is the corresponding unit vector. $R_{k l_1 l_2 l_3}^{(t)}$ is obtained by feeding a set of radial features that embed the radial distance $r_{ji}$ using Bessel functions multiplied by a smooth polynomial cutoff  to an MLP. It can be further simplified:

\begin{equation}
A_{i,k l_1 m_1}^{(1)} = \sum_{j \in \mathcal{N}(i)} R_{k l_1}^{(1)}(r_{ji}) Y_{l_1}^{m_1}(\hat{r}_{ji}) W_{k z_j}^{(1)}.
\end{equation}

This simplified operation is much cheaper, making the computational cost of the first layer low.

The {key operation} of MACE is the efficient construction of higher order features from the $A_i^{(t)}$-features:

\begin{equation}
\begin{split}
B_{i,\eta_v k L M}^{(t)} = \sum_{lm} C_{\eta_v, lm}^{LM}
\prod_{\xi = 1}^\nu \sum_{k'} w_{k k' \xi}^{(t)} A_{i,k' l_\xi m_\xi}^{(t)}, \\
\quad lm = (l_1 m_1, \ldots, l_\nu m_\nu)
\end{split}
\end{equation}

where the coupling coefficients $C_{\eta_v, lm}^{LM}$ correspond to the generalised Clebsch-Gordan coefficients ensuring that $B_{i,\eta_v k L M}^{(t)}$ are $L$-equivariant, the weights $w_{k k' \xi}^{(t)}$ are mixing the channels ($k$) of $A_i^{(t)}$, and $\nu$ is a given correlation order.


The message $m_i^{(t)}$ can now be written as a linear expansion

\begin{equation}
m_{i,kLM}^{(t)} = \sum_{\nu} \sum_{\eta_v} W_{z_i k L \eta_v}^{(t)} B_{i,\eta_v k L M}^{(t)},
\end{equation}

where $W_{z_i k L \eta_v}^{(t)}$ is a learnable weight matrix that depends on the chemical element $z_i$ of the receiving atom and message symmetry $L$.

\paragraph{Update}
The update is a linear function of the message and the residual connection:

\begin{equation}
\begin{split}
h_{i,kLM}^{(t+1)} = U_t^{kL}(\sigma_i^{(t)}, m_i^{(t)}) =
\sum_{\tilde{k}} W_{kL,\tilde{k}LM}^{(t)} m_{i,\tilde{k}LM}^{(t)} \\
+ \sum_{\tilde{k}} W_{z_i kL \tilde{k}}^{(t)} h_{i,\tilde{k}L M'}^{(t)}
\end{split}
\end{equation}

\paragraph{Readout}
In the readout phase, the invariant part of the node features is mapped to a hierarchical decomposition of site energies via readout functions:

\begin{equation}
E_i = E_i^{(0)} + E_i^{(1)} + \ldots + E_i^{(T)}, \quad \end{equation}
where
\begin{equation} \quad
E_i^{(t)} = \mathcal{R}_t(h_i^{(t)}) = 
\begin{cases}
\sum_{\tilde{k}} W_{\text{readout},\tilde{k}}^{(t)} h_{i,\tilde{k}00}^{(t)} & \text{if } t < T \\\\
\text{MLP}_{\text{readout}} \left( \{ h_{i,k00}^{(t)} \}_k \right) & \text{if } t = T
\end{cases}
\end{equation}

The readout only depends on the invariant features $h_{i,k00}^{(t)}$ to ensure that the site energy contributions $E_i^{(t)}$ are invariant as well.

